\documentclass[11pt, a4paper, onecolumn,  copyright, goog]{google}

\usepackage[authoryear, sort&compress, round]{natbib}

\usepackage{array}
\newcolumntype{L}[1]{>{\raggedright\let\newline\\\arraybackslash\hspace{0pt}}m{#1}}
\newcolumntype{C}[1]{>{\centering\let\newline  \\\arraybackslash\hspace{0pt}}m{#1}}
\newcolumntype{R}[1]{>{\raggedleft\let\newline \\\arraybackslash\hspace{0pt}}m{#1}}

\usepackage{geometry}
\usepackage{amsmath}
\usepackage{amssymb}
\usepackage{graphicx}
\usepackage{listings}
\usepackage{booktabs}
\usepackage{xcolor}
\usepackage{hyperref}
\usepackage{authblk}
\usepackage{float}
\usepackage{caption}
\usepackage{subcaption}

\hypersetup{
    colorlinks=true,
    linkcolor=blue!60!black,
    citecolor=green!50!black,
    urlcolor=blue!60!black
}

\newcommand{\Wm}{\,\text{W/m}^2}
\newcommand{\degr}{^{\circ}}
\newcommand{\mW}{\ensuremath{\,\mathrm{mW}}}
\newcommand{\mWh}{\ensuremath{\,\mathrm{mW}{\cdot}\mathrm{h}}}

\usepackage{listings}
\usepackage{xcolor} 

\uselogo{}

\title{Optimized Three-Dimensional Photovoltaic Structures with LLM guided Tree Search}

\author[1,2]{Michael P. Brenner}
\author[1]{Lizzie Dorfman}
\author[1]{John C. Platt}
\affil[1]{Google Research}
\affil[2]{School of Engineering and Applied Sciences, Harvard University}
\date{\today}
\begin{abstract}
We present a case study for how AI coding systems can be used to generate novel scientific hypotheses.
We combine a generic coding agent (Google's AntiGravity) with  an LLM-driven tree search algorithm (Empirical Research Assistance / ERA) to autonomously generate high-efficiency three-dimensional photovoltaic (3DPV) structures that overcome losses limiting flat solar panels at mid-latitudes. These structures operate by presenting
favorable angles to the sun throughout the day, and for illustrative purposes we focus on optimizing performance for a single solar day. Our workflow begins by using AntiGravity to reproduce calculations \cite{bernardi2012solar} showing that 3DPV can have energy densities much higher than stationary flat PV panels. We use these initial designs as the starting point for large scale tree search, where we seek improved solutions and score them for their diurnal yield. The initial tree search leads to nominally more efficient solutions, yet they are caused by algorithmic reward hacking,  arising from non-physical design features such as structurally levitating disconnected tiers and exploitations of the discretizations in the optics solver. To counteract this, we develop a workflow where the coding agent iteratively patches the physics engine with constraints to eliminate reward hacking.  With reward-hacking eliminated, ERA discovers 
a series of designs with various constraints and improved performance, including optimal designs with different fixed collector areas, optimizing zenith tracking and avoiding self shadowing.
 Combining coding agents with tree search (ERA) provides a powerful platform for scientific discovery, for problems whose solutions can be empirically evaluated with a score function.
\end{abstract}
\begin{document}
\maketitle

\section{Introduction}
\label{sec:intro}
Agentic AI systems have enormous potential for carrying out sophisticated analyses of scientific problems, accelerating the exploration of unconventional ideas. To demonstrate this capability, this paper presents a case study illustrating how we combined Empirical Research Assistance (ERA)—an autonomous LLM-driven tree search algorithm—with a coding agent (Google's Antigravity) to tackle a complex optimization problem in three-dimensional photovoltaic (3DPV) design. 

Conventional flat photovoltaic installations are the dominant method for solar energy collection on residential or commercial buildings, yet their performance at mid-to-high latitudes suffers from a fundamental geometric limitation: the cosine projection of direct sunlight onto the panel surface. At 7:00 AM on the summer solstice at latitude $42.36^{\circ}$ N (Boston, Massachusetts), the solar elevation is approximately $20^{\circ}$, yielding $\cos(70^{\circ}) \approx 0.34$. This means that a flat horizontal panel captures barely one-third of the available direct normal irradiance. Averaged over the full day, this cosine loss concentrates the majority of energy generation into a narrow $\sim$4-hour window around solar noon, leaving early-morning and late-afternoon flux largely unharvested. Utility photovoltaics can use tracking systems that adapt to the Sun's position and improve power performance by 1.2-1.8x\cite{bernardi2012solar}  but such tracking systems are impractical for residential and commercial owners.

\cite{bernardi2012solar} made the intriguing suggestion that three dimensional
 photovoltaic (3DPV) structures could address the geometric limitation by arranging solar cells on non-coplanar surfaces that present favorable angles to the sun throughout the day. Through numerical calculations and experiments,  \cite{bernardi2012solar}  demonstrated that 3D structures can generate several times more energy per base area than flat panels, depending on the detailed geometry and height-to-footprint ratio. The study assumed that the 3DPV would be built on a flat roof with fixed area: the goal was to maximizes the energy generated under that constraint. The study also assumed only direct irradiance would generate electricity.

Inspired by their study, we explore whether modern coding agents can efficiently generate improved designs and hypotheses for 3DPV structures.
Our workflow begins by providing a coding agent (Google's Antigravity) with the PDF of \cite{bernardi2012solar} \cite{bernardi2012solar} and asking it to reproduce the main results. The  system autonomously identified the GitHub repository used to carry out the original calculations, resolved legacy dependencies, and replicated the baseline energy limits of both the planar arrays and the human-designed  "open cube" structure  reported in the original 2012 study. We then tasked ERA to maximize the total energy yield of the 3DPV structure.
The system responded by discovering how to exploit the rendering code, finding ways of increasing yield that violate the laws of physics. Examples include a solution of levitating, disconnected solar arrays that maximize ground-view sunlight capture without incurring self-shading penalties; pushing structures outside the geographical bounding box to evade shadows entirely, and exploiting floating-point discretization limits.  To contain these exploits, we developed a workflow  by using the coding agent to 
iteratively patch the physics code and the scoring function. When suitably constrained, the system was able to find a series
of valid 3dPV architectures with superior performance, obeying different sets of design constraints. Overall this system demonstrates that the combination of coding agents with tree search is a powerful tool for efficiently discovering novel hypotheses.

\section{Mathematical Model}
Bernardi {\sl et. al.}'s original simulation code uses classical thin-film Fresnel optics to compute transmission through a planar encapsulant and traces shadowing between panels at sub-cell resolution.   We build our analysis from the author's Github repo with the original code\footnote{See https://github.com/feranick/3dpv.}.
For definiteness, they focus their calculations on a single date June~21, 2011 in Boston, MA, at coordinates
$(42.36\degr\text{N},\, 71.09\degr\text{W})$.  
The algorithm computes sun's position  using the Solar Position Algorithm
(SPA)~\cite{reda2004solar}, which yields the solar zenith angle $\theta_z$
and azimuth $\phi_s$ as functions of time, date, and geographic
coordinates. On this day in this location, the zenith angle ranges
from $90\degr$ (sunrise/sunset) to approximately $19\degr$ (solar noon).  Our analysis of the code using AntiGravity identifies several small errors that we corrected in our implementation: as an example, Snell's transmission angle calculations on secondary bounces occurred before updating the refractive index ratio to match the targeted surface, resulting in incorrect polarization parameters. We also eliminated the ability of the single-sided panels to absorb direct light.

Within our model, a candidate 3DPV structure is composed of triangular
solar panels embedded in a $20 \times 20 \times 10\,$m bounding box
above a flat rooftop. Each triangle $T_k$ is defined by three vertices
$\mathbf{v}_{k,1}, \mathbf{v}_{k,2}, \mathbf{v}_{k,3} \in \mathbb{R}^3$
with $0 \leq x,y \leq 20\,$m and $0 \leq z \leq 10\,$m. For definiteness, we constrain the  total
panel surface area as follows:
\begin{equation}
  A_{\text{total}} = \sum_{k=1}^{N} \frac{1}{2}
  \left\| (\mathbf{v}_{k,2} - \mathbf{v}_{k,1}) \times
  (\mathbf{v}_{k,3} - \mathbf{v}_{k,1}) \right\| \leq 800\,\text{m}^2
  \label{eq:area_constraint}
\end{equation}

For ray tracking each panel
is subdivided into sub-cells. At each time
step (every 6~minutes from sunrise to sunset), the simulation:
\begin{enumerate}
  \item Computes the sun direction $\hat{\mathbf{s}}(\theta_z, \phi_s)$.
  \item For each sub-cell centroid $\mathbf{p}$, traces a ray toward the
        sun and tests for intersection with all other panels to determine
        the shadow state.
  \item Computes the incidence angle $\theta_i$ between $\hat{\mathbf{s}}$
        and the panel's outward normal $\hat{\mathbf{n}}_k$.
  \item Calculates the transmitted (absorbed) and reflected power using
        the model-specific optics (see below).
  \item Traces reflected rays for secondary absorption (one bounce).
\end{enumerate}

The daily energy is then obtained by trapezoidal integration over all time
steps:
\begin{equation}
  E = \int_0^{T} P(t)\,dt \approx
  \sum_{i=0}^{M-1} \frac{P(t_i) + P(t_{i+1})}{2}\,(t_{i+1} - t_i),
  \label{eq:energy_integral}
\end{equation}
where $P(t_i)$ is the total instantaneous power across all panels at
time $t_i$. Note that this calculation focuses on the ideal theoretical optical limit and doesn't account for string-level electrical mismatch losses.

At each time step, the simulation traces direct optical vectors toward the absolute solar alignment state. We treat each panel
a single-sided collector behind a flat transparent cover with refractive index $n_2 = 2.2$ (relative to air, $n_1 = 1.0$), matching the  parameters of \cite{bernardi2012solar}. 
Equation~\eqref{eq:legacy_irradiance} dictates the direct normal solar irradiance follows the true Air Mass attenuated solar variable:
\begin{equation}
  I_{\text{legacy}} = 1488\Wm \times 0.7^{AM}
  \label{eq:legacy_irradiance}
\end{equation}
where $AM = (1/\cos\theta_z)^{0.678}$.
When unpolarized light strikes the interface of the panel at incidence angle $\theta_i$, Snell's law dictates the transmitted angle $\theta_t$ obeys $n_1 \sin\theta_i = n_2 \sin\theta_t$, whereas the Fresnel equations give the magnitude of reflectance $R_{s,p}$ for the polarizations $(s,p)$ and transmittance $T$ as

\begin{align}
  R_s &= \left( \frac{n_1 \cos\theta_i - n_2 \cos\theta_t}{n_1 \cos\theta_i + n_2 \cos\theta_t} \right)^2 \label{eq:rs_model} \\
  R_p &= \left( \frac{n_1 \cos\theta_t - n_2 \cos\theta_i}{n_1 \cos\theta_t + n_2 \cos\theta_i} \right)^2 \label{eq:rp_model} \\
  R &= \frac{R_s + R_p}{2} \label{eq:r_avg_model} \\
  T(\theta_i) &= 1 - R \label{eq:t_transmission}
\end{align}
Putting this together we can compute the unshadowed collected power as
\begin{equation}
  P_k^{(\text{legacy})} = T(\theta_i)\,I_{\text{legacy}}\,\cos\theta_i \cdot A_{\text{net},k} \cdot \eta
  \label{eq:legacy_power}
\end{equation}
where $\eta = 0.12$ fixes total static legacy cell efficiencies natively.

\begin{figure}[htbp]
    \centering
    \includegraphics[width=0.8\linewidth]{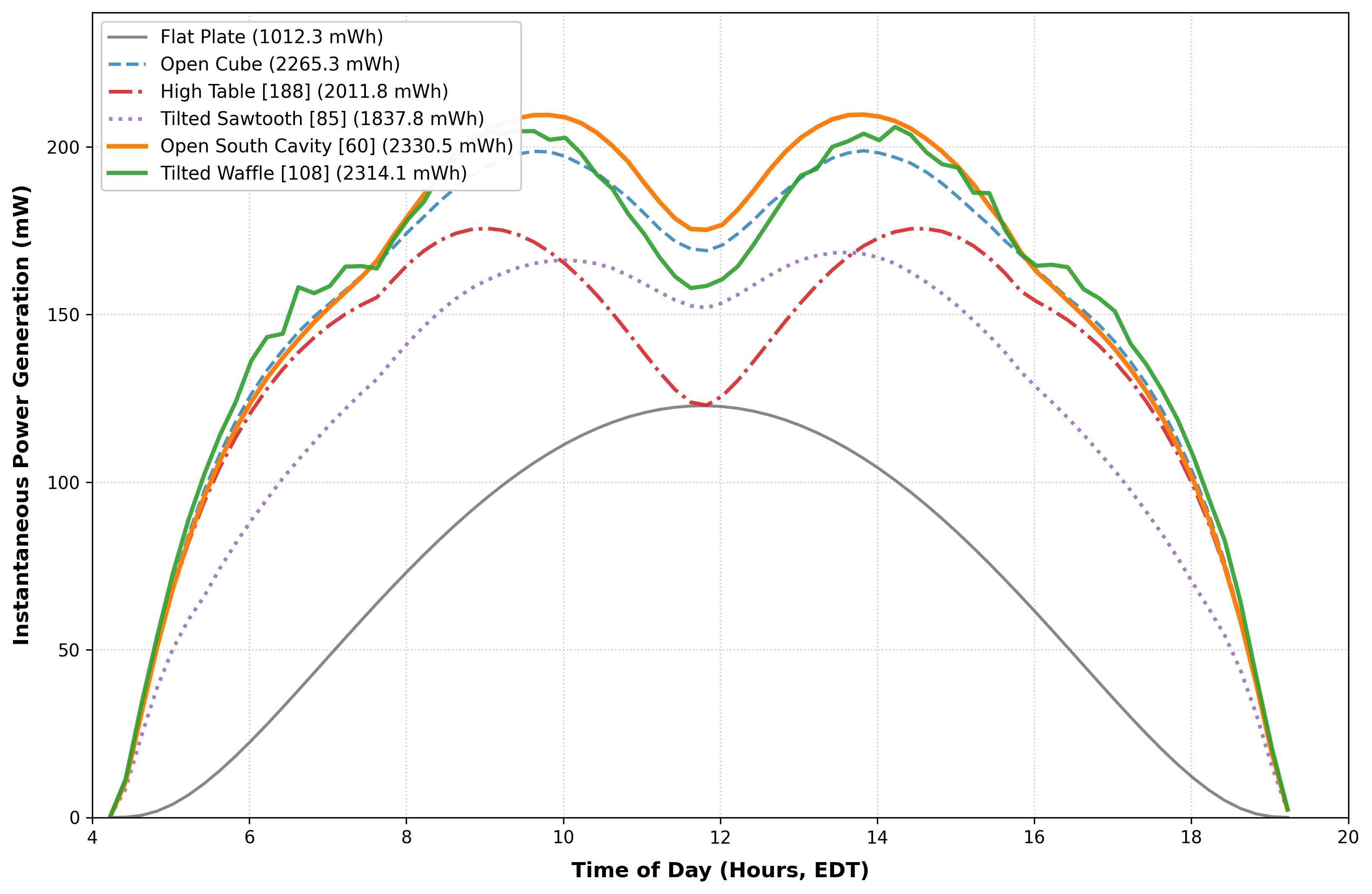}
    \caption{Unified comparative light curves across baseline and optimized geometries under True Legacy physics, including the flat panels and open cube design from \cite{bernardi2012solar}, as well as the optimal designs we develop in this paper: these include High Table, Tilted Sawtooth, Open South Cavity, and Tilted Waffle.}
    \label{fig:unified_light_curves}
\end{figure}

\cite{bernardi2012solar} used this formalism to compute light curves of power versus time for different designs. Figure~\ref{fig:unified_light_curves} reproduces these results, comparing the response of the flat plate to that of the cube.

\section{Empirical Research Assistance Tree Search}
\label{sec:legacy_exploits}
Our goal is to find designs that beat these initial ideas, using an LLM-guided tree search. 
Empirical Research Assistance (ERA) is an LLM-driven tree search system designed to optimize code-based scientific problems~\cite{aygun_ai_2025}.
Unlike classical evolutionary algorithms that
mutate parameter vectors, ERA operates directly on \emph{source code}:
a large language model reads a candidate program, analyzes its
performance, and proposes a modified version aimed at improving the
objective. This makes ERA naturally suited to combinatorial design
problems where the search space is defined by the geometry-generating
code rather than a fixed-dimensional parameter vector.  

We set up an ERA prompt that sets up the problem and asks the LLM to generate the geometry of the design.  The design is then scored by calculating total daily solar yield using  the physics simulator.  In particular, we prompt the LLM to implement the following function:

\begin{lstlisting}
def generate_geometry(params: dict) -> str:
    """
    Generates the string contents defining a 3D solar cell mesh geometry.
    
    Args:
        params: Dictionary of configurable geometric hyperparameters to tune.
        
    Returns:
        A formatted string complying with the solar3d custom shape descriptor format.
        Format rule: 
        - Return ONLY a sequence of floating-point numbers.
        - DO NOT prefix the geometry string with the number of triangles! The evaluator will automatically detect the triangle count.
        - Each triangle is defined by exactly 9 floats space-separated defining the 3 vertices (x,y,z): "x1 y1 z1 x2 y2 z2 x3 y3 z3"
        - The triangles can be separated by spaces or newlines.
        - Important Security Rule: ALL coordinates MUST strictly reside within the bounding box: X [0.0, 20.0], Y [0.0, 20.0], Z [0.0, 20.0]. The evaluator will strictly enforce this and return 0.0 kWh if any coordinates are violated.
    """
    pass
\end{lstlisting}

We score this function using the same physics simulator as \cite{bernardi2012solar}, with minor errors corrected as outlined above. Additionally, when we carry out our optimization we impose constraints: a natural constraint is to fix the total area of the solar cells as this correlates strongly with cost.  We consider three different area constraints: allowing the 3DPV to have an area of 3 or 5 $\times$ that of the initial flat collector, as well as a more unconstrained design allowing an area of up to 20$\times$ the initial collector. 
We emphasize that a strength of this method  is its flexibility:  it is straightforward to modify the simulator for other models, such as angle-dependent reflectance from nanotextured coatings, diffuse irradiance and so forth, and equally straightforward to modify the constraints.  

\section{Discovering and Attenuating Algorithmic Exploits}
\label{sec:legacy_exploits}

Our initial LLM guided tree search runs produced extremely high scores, but did so by identifying strategies that  violate the laws of physics. The exploits include structurally levitating disconnected tiers and exploiting the discreteness of the optics simulator.  To address this we adopted a workflow whereby we iteratively run tree search with a given scoring function; analyze the resulting solutions and determine whether they violate physical constraints, and then update the scoring function or potentially the physics simulator to address these constraints (Figure~\ref{fig:adversarial_workflow_c}). 
We used Antigravity to  uncover these violations and suggest updates to the scoring function.  In what follows we describe each of the exploits in turn and the solution.

\begin{figure}[htbp]
\centering
\fbox{
  \begin{minipage}{0.7\columnwidth}
    \vspace{0.2em}
    \centering
    \textbf{\large Workflow for Eliminating Reward Hacking  }
    \vspace{0.5em}
    \hrule
    \vspace{0.5em}
    \begin{enumerate}
      \item \textbf{ERA proposes geometries.} ERA explores non-intuitive, complex structural arrangements specifically optimized to challenge integration edge cases across the continuous domain.
      \item \textbf{ Fresnel Simulator evaluates bounds.} We simulate the physics for proposed  geometries to evaluate their performance.
      \item \textbf{Detect Reward Hacking.} Antigravity (together with human expert driving the search) examines outputs  to separate genuine  improvements from those that violate the laws of physics.
      \item \textbf{Antigravity Coding Agent writes Python patches.} Rewrite either the physics simulator or employ constraints to disallow reward hacking.  
      \item \textbf{Repeat.} Relaunch an ERA search, aiming to find novel solutions that are more realistic. 
    \end{enumerate}
    \vspace{0.2em}
  \end{minipage}
}
\caption{ An iterative framework for eliminating reward hacking, by improving scoring functions to find physically realistic, optimal solutions.}
\label{fig:adversarial_workflow_c}
\end{figure}

\subsection{Exploit 1: Levitating Structures}
\label{sec:levitating_structures}
\begin{figure*}[htbp]
    \centering
    \begin{subfigure}[b]{0.49\textwidth}
        \centering
        \includegraphics[width=\textwidth]{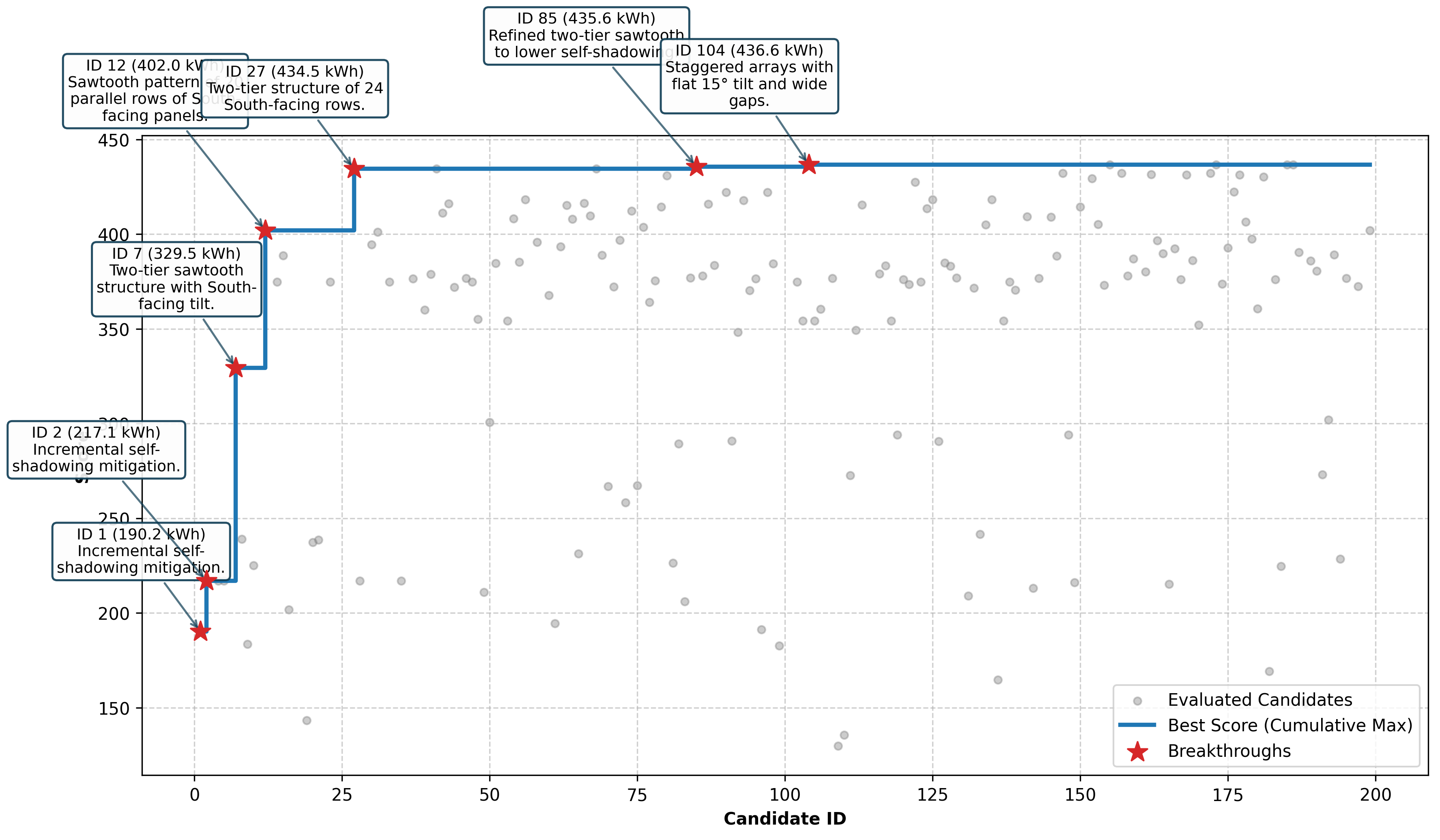}
        \label{fig:exploit1_search}
    \end{subfigure}
    \hfill
    \begin{subfigure}[b]{0.49\textwidth}
        \centering
        \includegraphics[width=\textwidth]{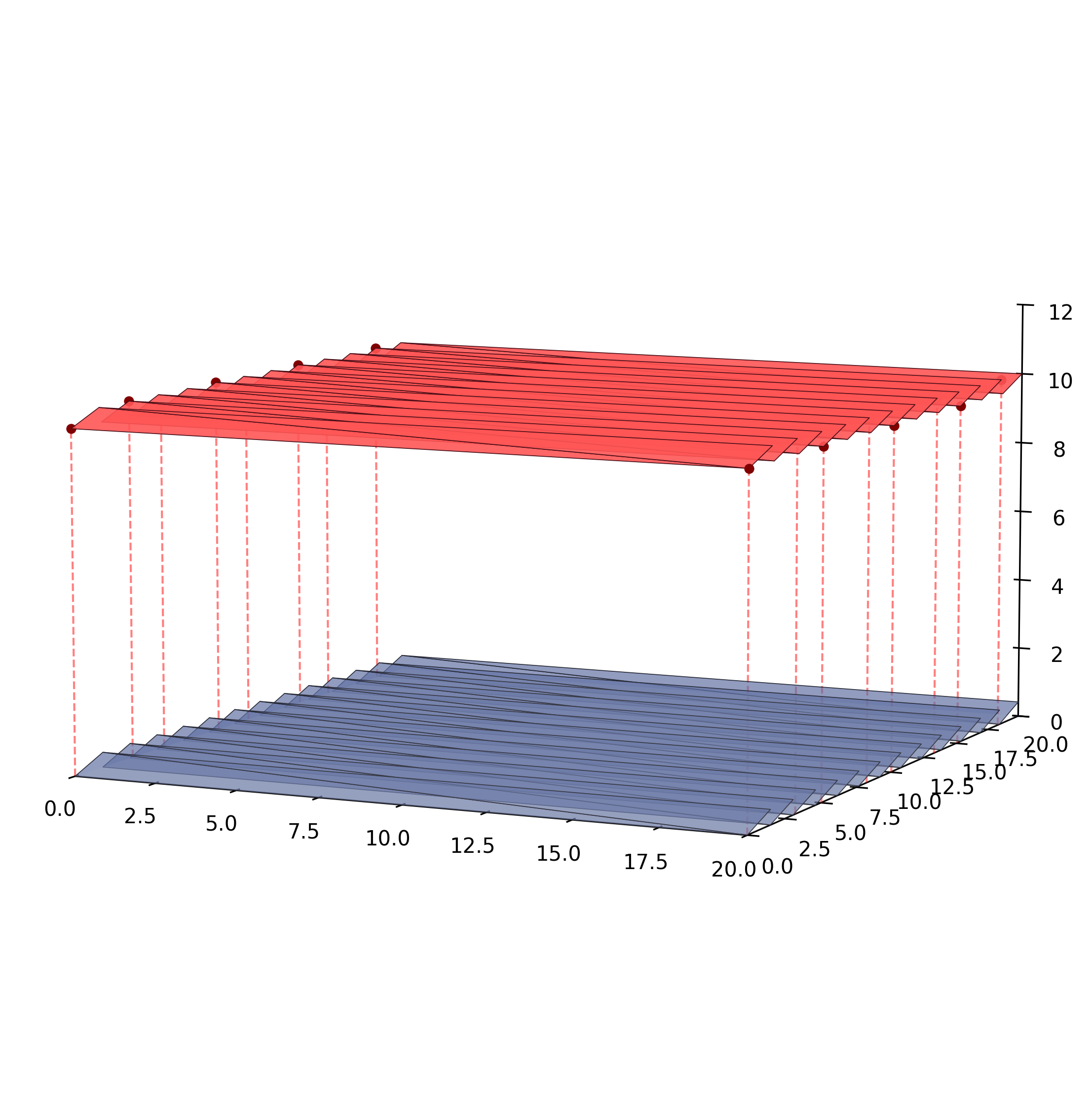}
        \label{fig:legacy_breakthrough_1}
    \end{subfigure}
    \caption{Milestone 1 exploit analysis: (a) Algorithmic search progression discovering the design that both increases efficiency but also violates  the laws of physics. (b) Optimal state attained by Candidate 104. Physical panels  levitate  completely unsupported in mid-air ($Z \approx 10.0$\,m) to  bypass shading interactions. }
    \label{fig:milestone1_combined}
\end{figure*}

\begin{figure}[ht]
    \centering
    \includegraphics[width=0.7\linewidth]{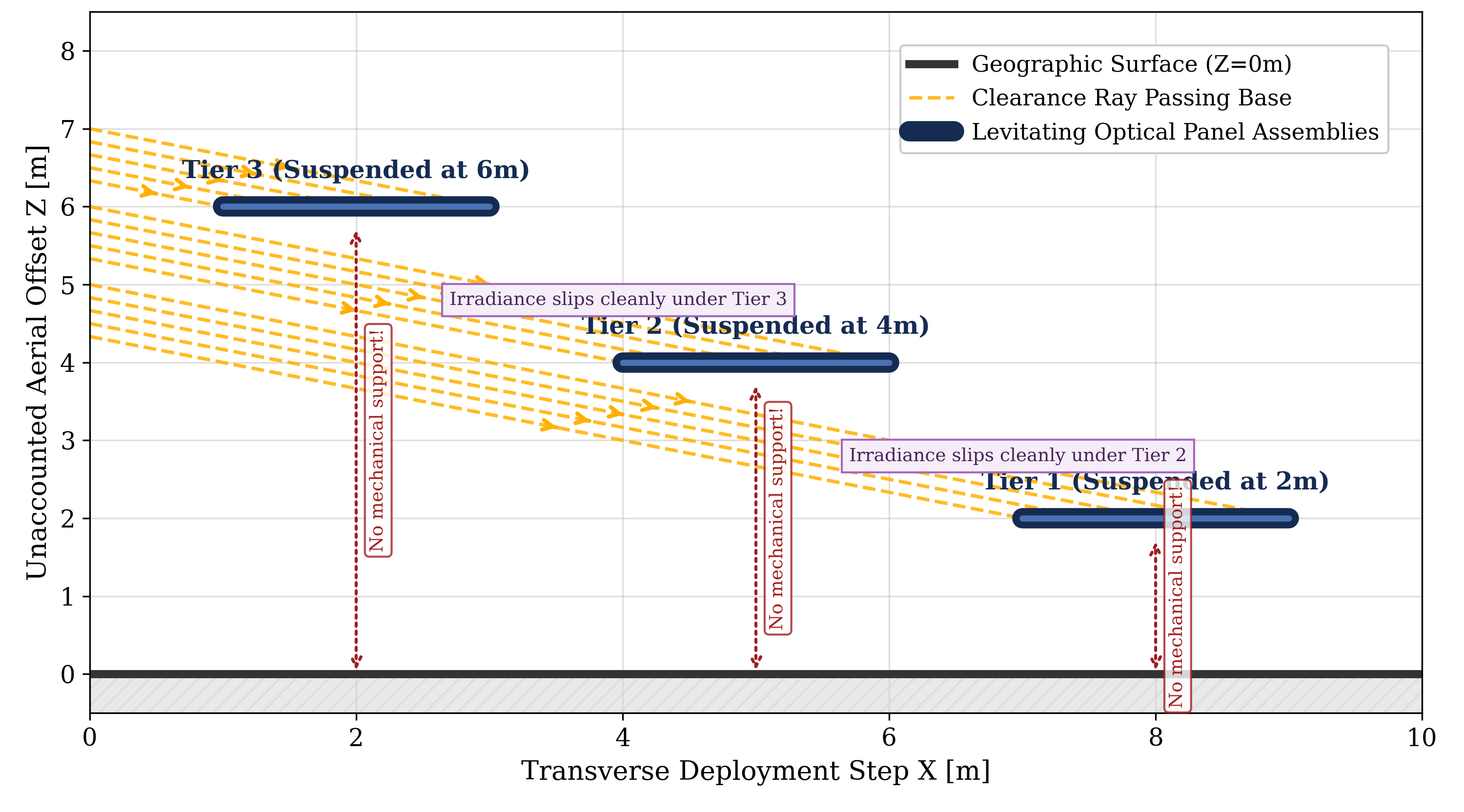}
    \caption{Dispersed light vectors slipping continuously below levitated surface plates.}
    \label{fig:exploit1_mech}
\end{figure}

Our initial search found excellent solutions about 4x more efficient than the flat plate design.  Yet,
these designs required  suspending several of the panels within the three dimensional volume without a physical anchor
(Figure~\ref{fig:milestone1_combined}).  The design increases efficiency by allowing dispersed light vectors to pass  below the levitated plates (Figure~\ref{fig:exploit1_mech}) for some range of angles, thus increasing overall collection efficiency.

 Antigravity identified this exploit, and proposed an update to the scoring function to eliminate it. The idea is to use a
 graph-theoretic Breadth-First Search (BFS) that dynamically scales and integer-hashes each vertex. The BFS algorithm seeds its traversal queue strictly with ground-anchored triangles  and traverses the array across shared vertices. If the final count of connected panels does not equal the total array count, the structure contains orphaned or levitating components and is strongly penalized.

\subsection{Exploit 2: Exploiting discretization in the optics solver}
\label{sec:phase_sync}

With the geometric exploit patched, ERA again found  high-scoring solutions,  this time composed of submillimeter slivers and stacked overlapping panels).
Yet on closer examination this high score also arose due to an exploit: 
This time ERA exploited the discrete nature of the numerical ray-tracer, and generated degenerate sub-millimeter slivers and stacked overlapping panels to intercept light multiple times without registering physical occlusions.  The optimizer discovered that physical ray-tracing happens at discrete intervals; creating compressed sub-millimeter inter-panel gaps caused the simulator's floating-point checking mechanism to fail to register solid boundaries, thereby allowing optical rays to bypass solid material without rendering shadows.
 Modulating structural boundaries onto microscopic elements enabled unattenuated light passages directly down beneath the intersection detection updates.

\section{The Optimal Designs}
\label{sec:architectural_geometry}
Having  fixed attempts to reward hack  the physics simulator, we transitioned to using ERA to discover new designs.
We carried out an ERA search using different sets of constraint: restricting the  area of the surfaces to at most $3\times$ that of a flat plate; restricting the surface area to $5\times$ the flat plate. The latter corresponds to the Open Cube design from \cite{bernardi2012solar}. Finally we considered a less constrained optimization that let the area go up to $20\times$ the flat plate.

\subsection{Optimizing with a $3\times$ Constraint}
\label{subsec:performance_mechanics}

\begin{figure*}[htbp]
    \centering
    \begin{subfigure}[b]{0.32\textwidth}
        \centering
        \includegraphics[width=\textwidth]{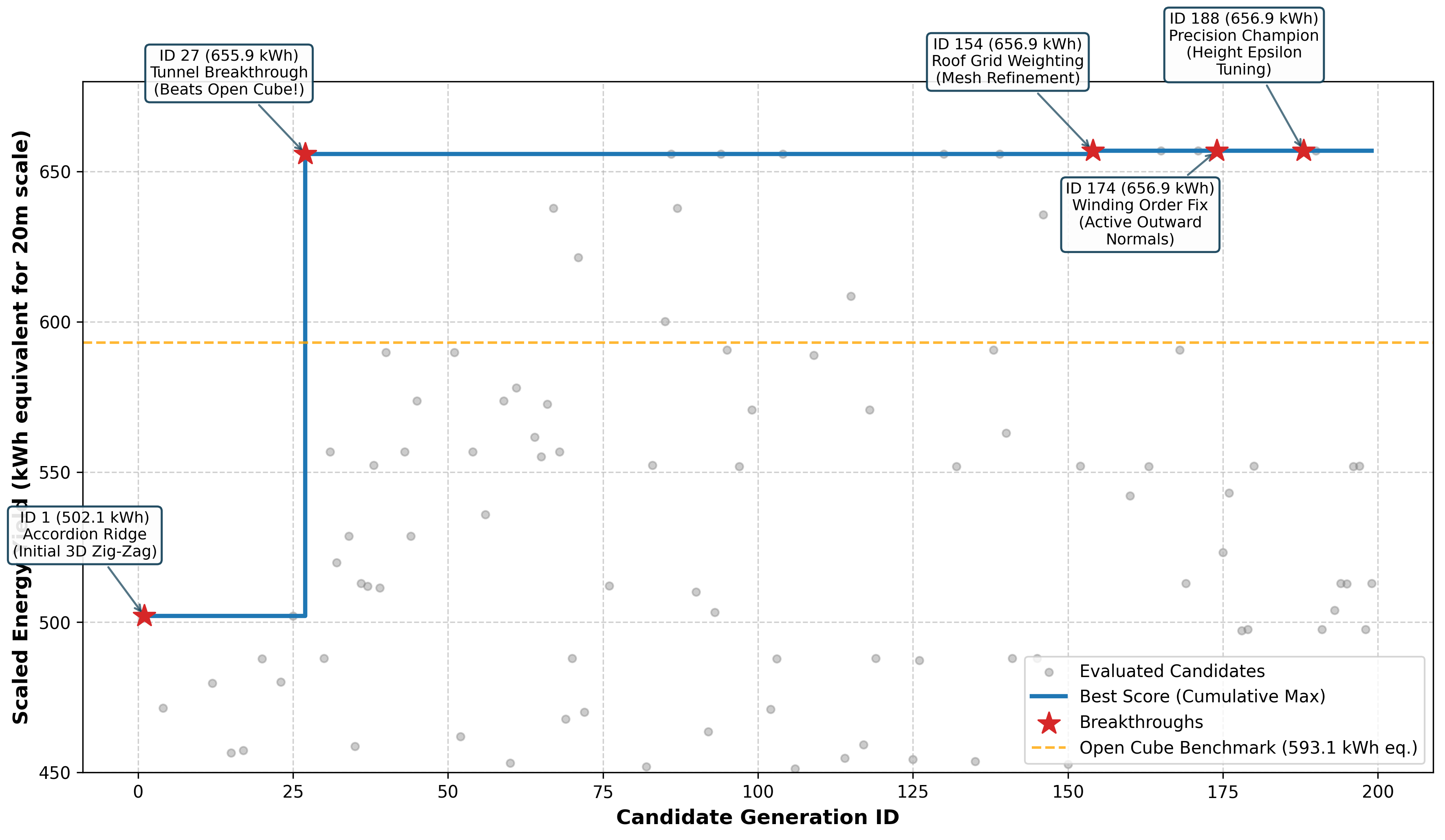}
        \caption{}
        \label{fig:breakthrough_learning_curve}
    \end{subfigure}
    \hfill
    \begin{subfigure}[b]{0.32\textwidth}
        \centering
        \includegraphics[width=\textwidth]{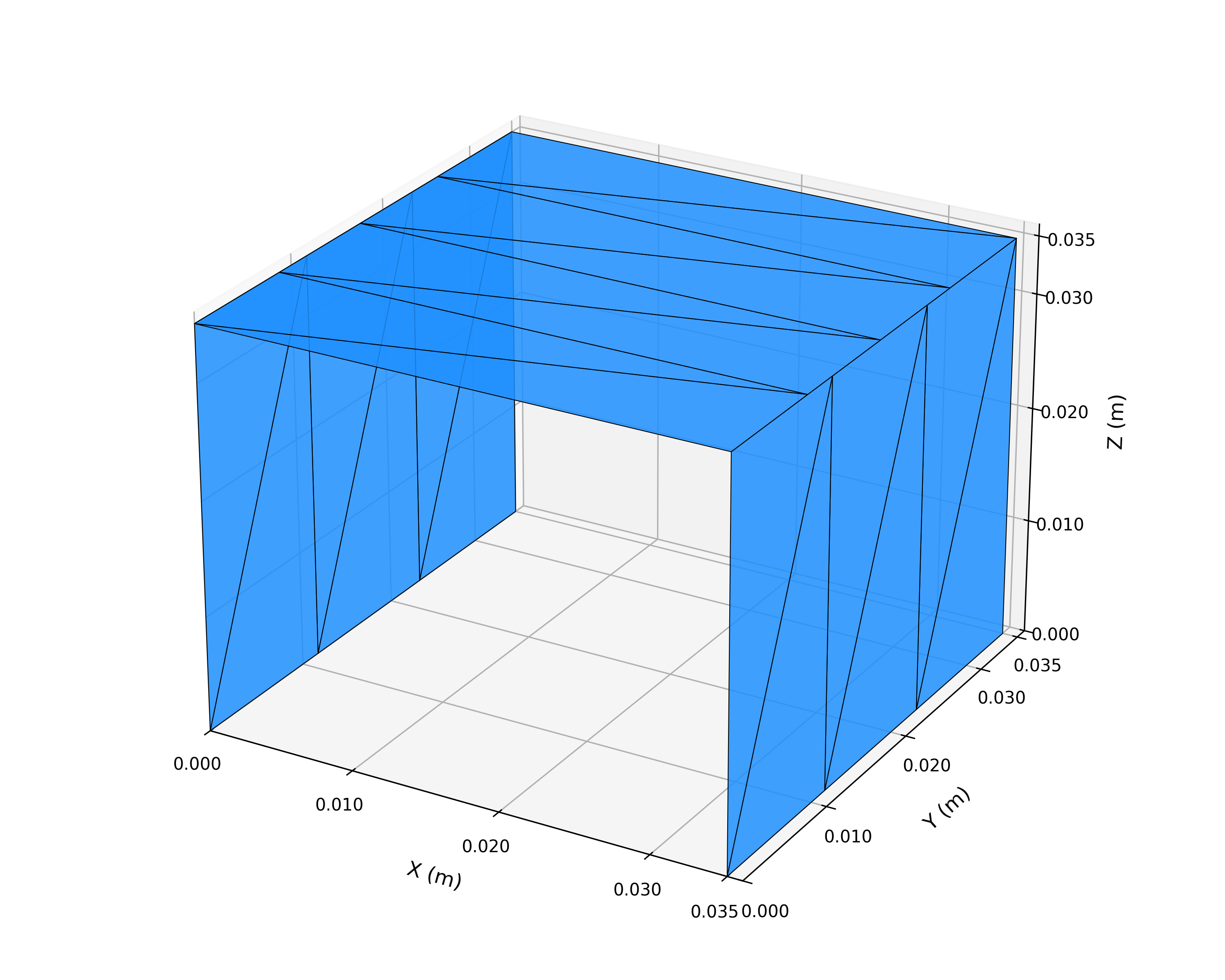}
        \caption{}
        \label{fig:cad_hero_188}
    \end{subfigure}
    \hfill
    \begin{subfigure}[b]{0.32\textwidth}
        \centering
        \includegraphics[width=\textwidth]{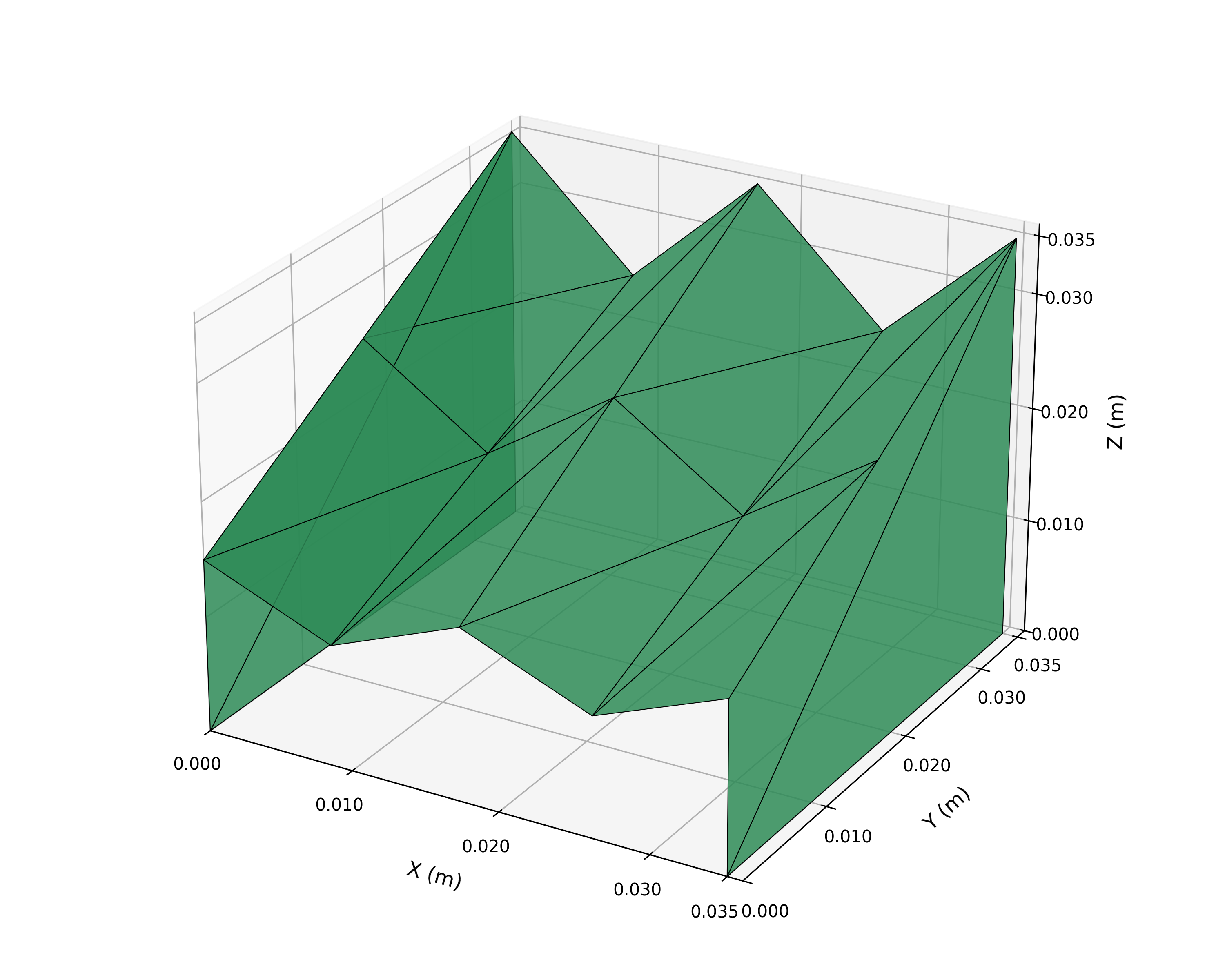}
        \caption{}
        \label{fig:cad_hero_85}
    \end{subfigure}
    \caption{Structural optimization renderings under strict 3$\times$ footprint area constraints: (a) Evolutionary learning curve showing the optimization breakthroughs leading to the best candidate (Number 188). (b) Structure of the best candidate, a High Table configuration ($2011.8\mWh$). (c) An intermediate best candidate (Number 85) with a M-Shape/Tilted Sawtooth configuration ($1837.8\mWh$).}
    \label{fig:cad_hero_render}
\end{figure*}

We begin by carrying out the ERA tree search assuming  the total area of the active collectors must not exceed three times the area of the flat base footprint ($\le 3s^2$).  The progression of the optimization is shown in Figure~\ref{fig:breakthrough_learning_curve}, which illustrates the evolutionary learning curve. Over the course of the search, the algorithm achieved several sequential breakthroughs, iteratively refining the structure's geometry to maximize diurnal solar yield.

The best solution from the search occurred at node 188 out of 200 nodes, an optimized 20-triangle ``High-Table'' architecture, arranging its surfaces into parallel East and West vertical walls, connected by a horizontal top roof. Crucially, the design leaves the North and South faces open,  satisfying the total surface area constraint and allowing light to penetrate. As the search proceeds a number of creative designs were invented. As one example,  an intermediate best candidate (number 85) has the shape of a Tilted Sawtooth. This solution adapts to the summer solstice solar track by using an asymmetric M-shaped corrugated roof tilted towards the South (Figure~\ref{fig:cad_hero_render}) to maximize oblique direct photon capture while retaining vertical East-West walls.  .

Under the patched physics model, Candidate 188 yields $2011.8\mWh$ of total daily energy and $175.6\mW$ of peak power, whereas Candidate 85 achieves $1837.8\mWh$ of energy and $168.5\mW$ of peak power. The diurnal power curves for  compared to the Flat Panel and Open Cube geometries are detailed in Figure~\ref{fig:unified_light_curves}.
In comparison, the standard Open Cube baseline---with five fully active faces (bottom floor + 4 vertical walls)---has a total surface area of $5s^2$.  Despite this material deficit, Candidate 188 not only satisfies the strict area limitation but also captures $88.8\%$ of the absolute energy production of the unconstrained Open Cube ($2265.3\mWh$) while using $40\%$ less material. This highlights the exceptional efficiency of the High-Table topology in trapping and harvesting incoming solar radiation while respecting the required  bounds.

\subsection{Optimizing with a $5\times$ Constraint}
\label{sec:eval_5x}

\begin{figure*}[htbp]
    \centering
    \begin{subfigure}[b]{0.49\textwidth}
        \centering
        \includegraphics[width=\textwidth]{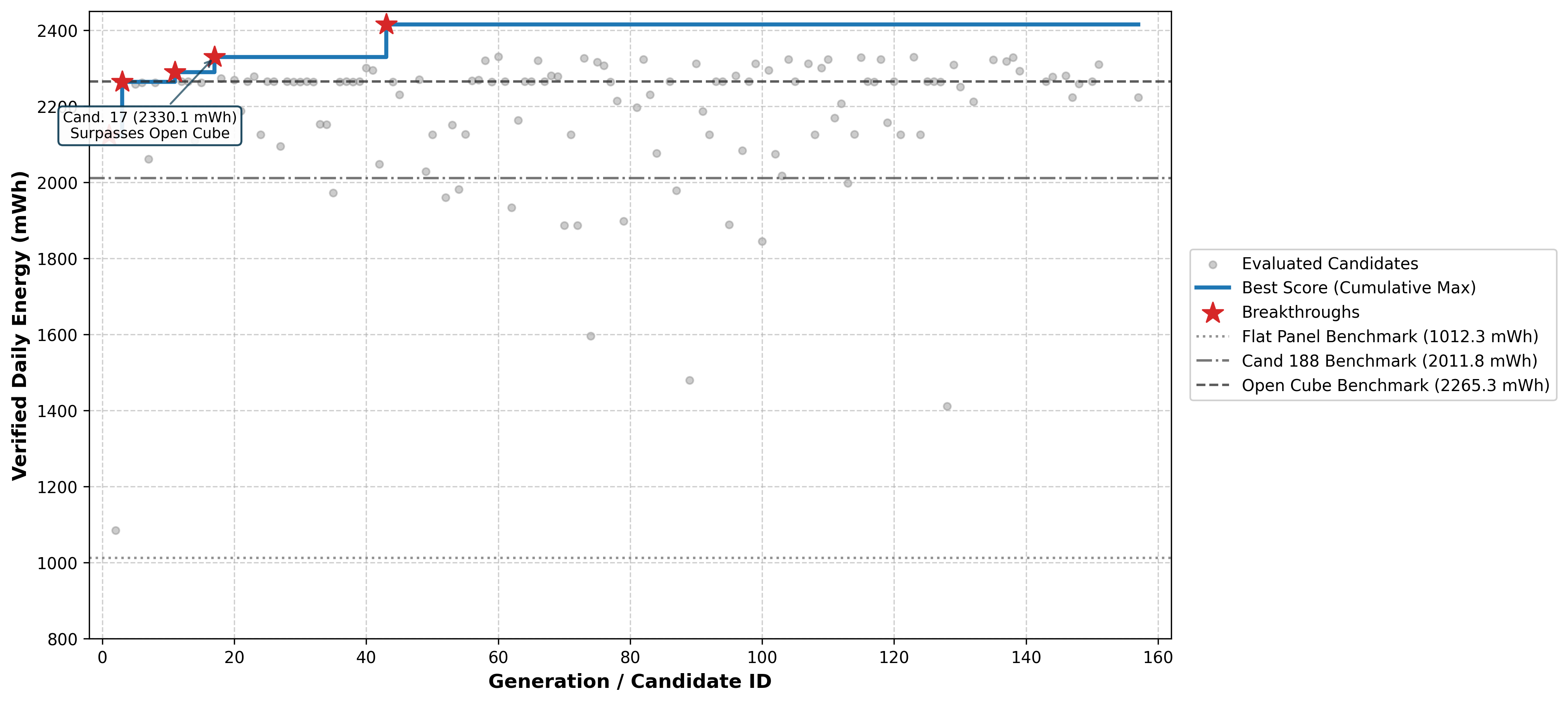}
        \caption{}
        \label{fig:training_trajectory_5x}
    \end{subfigure}
    \hfill
    \begin{subfigure}[b]{0.49\textwidth}
        \centering
        \includegraphics[width=\textwidth]{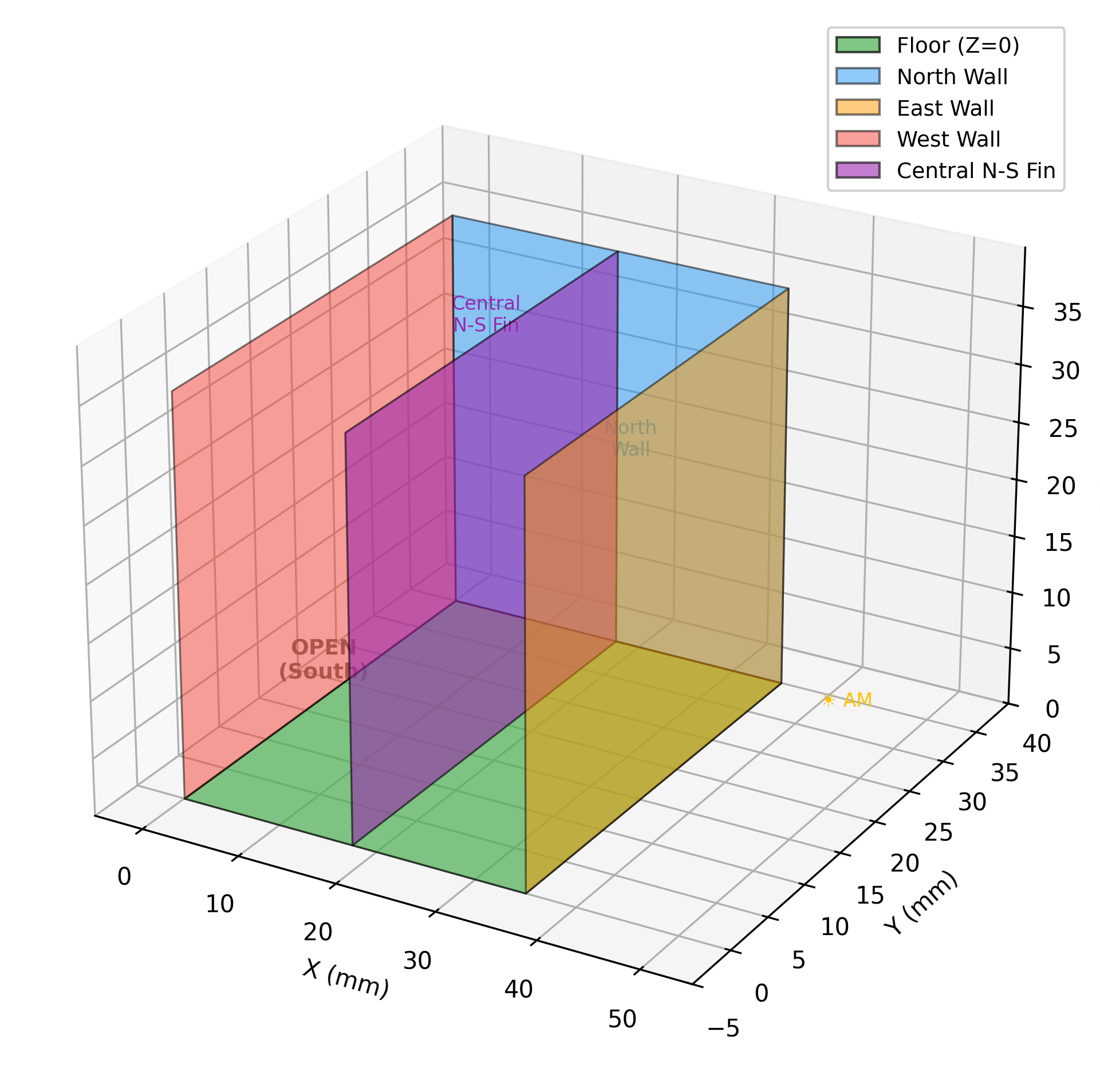}
        \caption{}
        \label{fig:geometry_cavity_fin_cand60}
    \end{subfigure}
    \caption{ERA search under the equal-material 5$\times$ surface limit: (a) Breakthrough plot. Over iterations the best candidate occurs at candidate 60.  (b) Three-dimensional layout of Candidate 60. The primary south-directed open cavity optimizes collection from the southern arc, whereas the internal central partition intercepts lower-elevation lateral rays.}
    \label{fig:cand60_combined}
\end{figure*}

To contrast against standard unconstrained Open Cubes using an identical material basis, we relaxed the constraint to a $5\times$ primary base surface constraint. The progression of the structured optimization campaign under this equal-material limit is shown in Figure~\ref{fig:training_trajectory_5x}. Evaluation across 200 continuous generations reveals the sequential geometric breakthroughs that allowed the algorithm to progressively explore the design space and enhance the daily yield.

This optimization yielded  a South-facing cavity structure featuring a central North-South oriented divider (presented in Figure~\ref{fig:geometry_cavity_fin_cand60}). This solution circumvents symmetrical East-West configurations to mitigate cosine fall-off effects at mid-day insolations. The structure's optimal static orientation utilizes the open south cavity for near-zenith direct flux while deploying the central internal divider to capture non-occluded direct irradiance at lower early morning and late afternoon solar elevation angles.   The solution achieves a maximum of $209.6\mW$ peak power alongside $2330.5\mWh$ of daily energy production, surpassing the performance of the Open Cube baseline by $+2.9\%$. A temporal comparison (Figure~\ref{fig:unified_light_curves}) demonstrates the relative generation advantage uniquely acquired during lower elevation periods. Table~\ref{tab:unified_output} summarizes the integrated performance across constraint conditions.

\subsection{Unconstraining the optimization: The Tilted Waffle}
\label{sec:eval_stepped_wedge}

While restricting the optimization to $3\times$ and $5\times$ footprint limits proves that ERA can efficiently arrange panels to match or slightly exceed human-designed volumetric configurations, we can push it further by expanding the material allowance. When we expanded the structural complexity to 50 active panels, allowing surface area scaling up to a $20\times$ footprint cap, ERA discovered a tilted waffle architecture.

\begin{figure*}[htbp]
    \centering
    \begin{subfigure}[b]{0.49\textwidth}
        \centering
        \includegraphics[width=\textwidth]{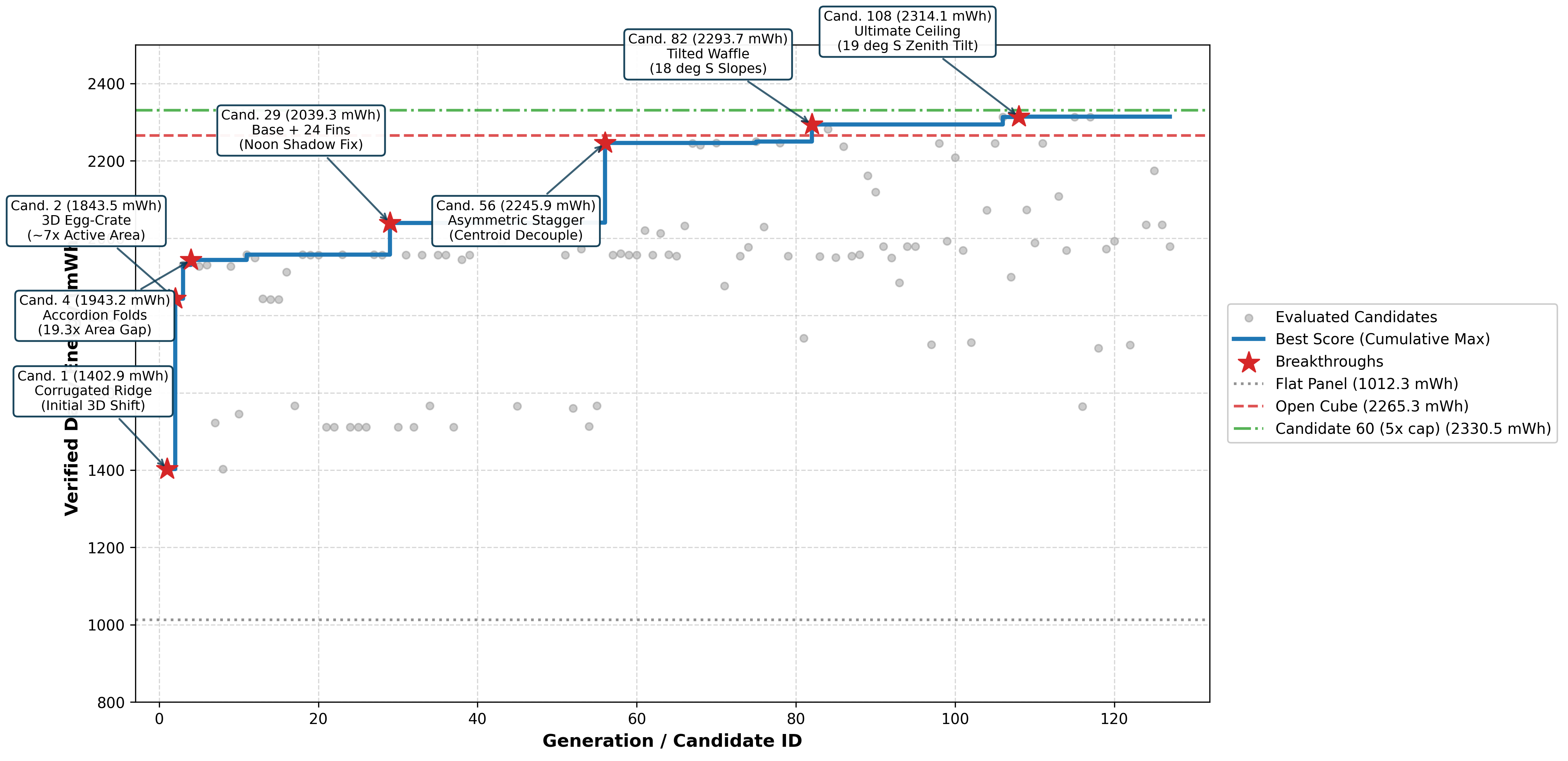}
        \caption{}
        \label{fig:breakthrough_50tri}
    \end{subfigure}
    \hfill
    \begin{subfigure}[b]{0.49\textwidth}
        \centering
        \includegraphics[width=\textwidth]{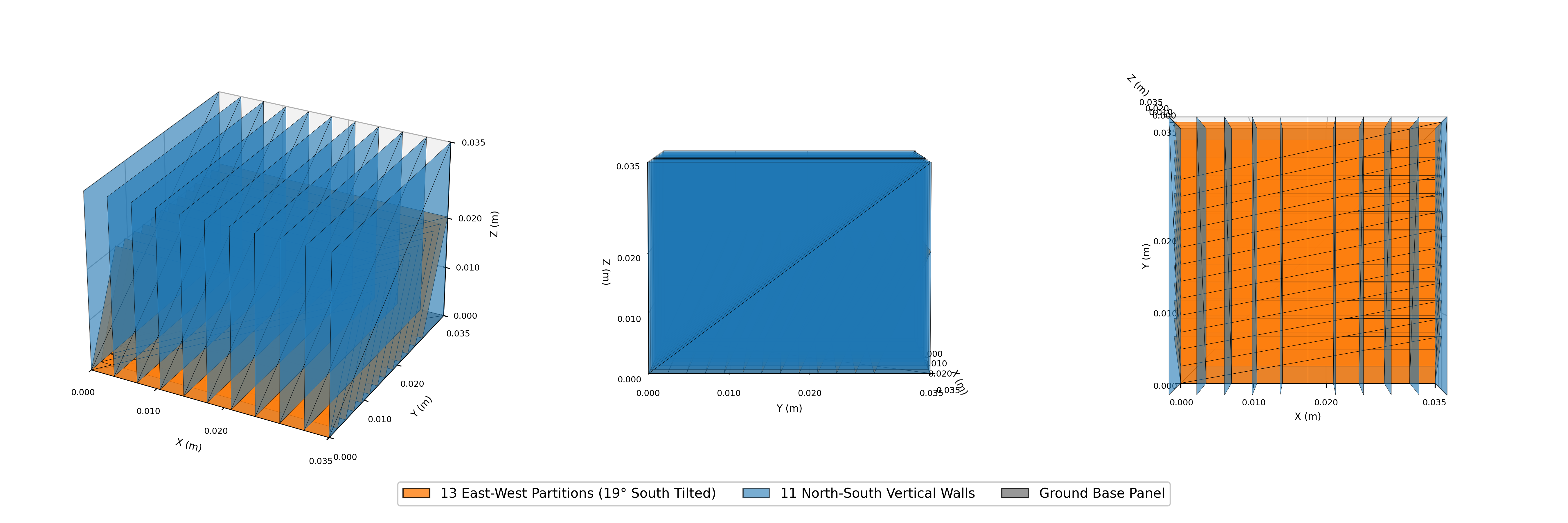}
        \caption{}
        \label{fig:cad_tilted_waffle}
    \end{subfigure}
    \caption{Less constrained ERA search, up to 20$\times$ footprint. The best candidate (number 108) has a ``Tilted Waffle'' structure: (a) Breakthrough plot, mapping the architectural transitions from accordion ridges to the ultimate Tilted Waffle. (b) Rendering of the optimal solution. North-South walls remain vertical while East-West partitions tilt South at $19\degr$ to track solar zenith.}
    \label{fig:cand108_combined}
\end{figure*}

The search explored a variety of designs, including  3D egg-crates and tight accordion folds, ultimately finding a geometry deploys 11 vertical North-South walls to trap lower-elevation East-West solar transit, and 13 East-West partitions tilted exactly $19\degr$ South (rendered in Figure~\ref{fig:cad_tilted_waffle}). The $19\degr$ South tilt matches the summer solstice culmination in Boston ($42.36\degr$\,N), tracking the solar zenith at noon. 

This solution captures a peak power of $205.9\mW$ and a total daily energy yield of $2314.1\mWh$ ($2.29\times$ relative gain),  the Open Cube. Yet, surprisingly, it does not beat the 20-panel Candidate 60  despite having quadruple the allowable surface area limit (Figure~\ref{fig:unified_light_curves}). This suggests a potential limit  for spatial shadow management. Scaling the structural grid to 50 panels forces inter-partition spacing down to ${\sim}1.75\,\text{mm}$, yielding extreme channel aspect ratios ($20:1$). This effectively collimates incoming rays and restricts light acceptance to a narrow window of only a few degrees, increasing the frequency of self-shadowing during both peak solar elevation and lateral morning/afternoon transitions.

\section{Conclusion}
\label{sec:conclusion}

Subjecting classical macroscopic physics simulators to open-ended algorithmic tree search illuminates both the presence of latent software vulnerabilities and the existence of hidden geometric optima. Our initial unconstrained operations reliably uncovered non-physical simulation states, including structurally levitating tiers, shadow-evasion outside geographic bounds, and spatial sub-millimeter tunneling. By formally shielding the computational framework with graph-theoretic structural checks, overlap penalties, and continuous geographic bounding, we successfully directed the LLM-driven optimizer away from reward hacking and back into properly grounded, physically viable topologies. The summary of results is shown in Table \ref{tab:unified_output}.

\begin{table*}[htbp]
\centering
\caption{Summary of performance outputs established against physical area limits.}
\label{tab:unified_output}
\begin{tabular}{llrr}
\toprule
Model / Structure & Constraint & Peak Power ($\mW$) & Daily Energy ($\mWh$) \\
\midrule
Baseline Flat Panel & $1\times$ base limit & $122.7$ & $1012.3$ \\
\midrule
Candidate 85 & $3\times$ active capacity & $168.5$ & $1837.8$ \\
Candidate 188 & $3\times$ active capacity & $175.6$ & $2011.8$ \\
\midrule
Open Cube & $5\times$ structural equivalent & $198.8$ & $2265.3$ \\
Cavity Fin & $5\times$ structural equivalent & $209.6$ & $2330.5$ \\
\midrule
Tilted Waffle & 50-cell, 20$\times$ cap & 205.9 & 2314.1 \\
\bottomrule
\end{tabular}
\end{table*}

The ERA framework isolated highly competitive structural niches for different sets of constraints. 
For each constraint, the algorithm mapped out a distinct hierarchy of architectures with high performance. At a restricted $3\times$ area cap, Candidate 188 demonstrates that static geometry alone can recover nearly $89\%$ of the energy of an unconstrained $5\times$ Open Cube while using $40\%$ less material. At the equal-material $5\times$ threshold, Candidate 60's asymmetric internal North-South baffle outperforms the human-designed Open Cube by $+2.9\%$, yielding $2330.5\mWh$. 

Finally, extending the optimization envelope to 50 panels and a $20\times$ area cap hinted at a  physical ceiling (the ``tilted waffle''). This design optimizes zenith capture via a customized $19\degr$ South tilt, but its daily energy yield ($2314.1\mWh$, $2.29\times$ baseline) and peak power ($205.9\mW$) shows negative marginal returns when compared against the less dense Candidate 60. This reveals a fundamental optical-geometric trade-off in dense 3DPV design: packing excessive active surfaces into a confined spatial volume achieves diminishing integrated yields due to compounding, unavoidable self-shadowing at all solar elevation angles. 
 Collectively, these results demonstrate that combining coding agents with LLM-driven tree search (ERA) provides a powerful platform for scientific discovery, capable of efficiently navigating complex physical constraints to propose novel, high-efficiency hypotheses for engineering and materials design.

\end{document}